# Highlights

- Implementation of detailed guidelines and a structured process to ensure high-quality annotations, focused on targeting language, from the crowd, experts, and ChatGPT to detect and analyze inappropriately targeting language in a comprehensive data set including conversations from banned subreddits

- Insights into toxic and non-toxic language

- Potential and limitations of OpenAI's GPT-3 (text-davinci-003) in the process of annotating inapropriately targeting language

- Examination of inter-annotator agreement challenges, including analysis of expert disagreements and performance comparison across the experts, crowd annotators and ChatGPT

# Understanding and Analyzing Inappropriately Targeting Language in Online Discourse: A Comparative Annotation Study

Baran Barbarestani[a,], Isa Maks[a], Piek Vossen[a]

[a]*Vrije Universiteit Amsterdam, De Boelelaan 1105, Amsterdam, 1081 HV, The Netherlands*

**Abstract**

Attention: This paper includes instances of hateful content for research purposes.

This paper introduces a method for detecting inappropriately targeting language in online conversations by integrating crowd and expert annotations with ChatGPT. We focus on English conversation threads from Reddit, examining comments that target individuals or groups. Our approach involves a comprehensive annotation framework that labels a diverse data set for various target categories and specific target words within the conversational context. We perform a comparative analysis of annotations from human experts, crowd annotators, and ChatGPT, revealing strengths and limitations of each method in recognizing both explicit hate speech and subtler discriminatory language. Our findings highlight the significant role of contextual factors in identifying hate speech and uncover new categories of targeting, such as social belief and body image. We also address the challenges and subjective judgments involved in annotation and the limitations of ChatGPT in grasping nuanced language. This study provides insights for improving automated content moderation strategies to enhance online safety and inclusivity.

## 1. Introduction

The study of combating internet hate speech using Natural Language Processing (NLP) has gained significant attention due to its scalability and efficiency in alleviat-

---

∗Corresponding author

*Email addresses:* b.barbarestani@vu.nl (Baran Barbarestani), isa.maks@vu.nl (Isa Maks), p.t.j.m.vossen@vu.nl (Piek Vossen)

ing the burden on human moderators [1]. Researchers are compiling extensive data sets from various sources to drive investigations in this field, comparing methodologies such as feature selection, machine learning (ML) techniques, and classification algorithms [2]. Despite this progress, less attention has been paid to how hate speech evolves within conversational contexts, a factor crucial for early detection and effective moderation on social media platforms. Understanding the early triggers of group targeting within online conversations is essential for identifying implicit hate speech and preventing it from escalating into explicit language. This paper aims to enhance our understanding of these dynamics, thereby contributing to the development of more effective content moderation strategies that foster inclusive and respectful online environments.

We introduce a novel dataset designed for benchmarking and analyzing targeting language in Reddit discussions. This dataset serves as a valuable resource for deepening our understanding of harmful online communication and evaluating the effectiveness of automated detection methods. Although the dataset is limited in size, it provides a foundational resource for future research and model evaluation. The study emphasizes advancing knowledge in the field and supporting the development of improved content moderation systems, rather than serving as a direct tuning tool. Our analysis includes annotations from crowd sources, experts, and OpenAI's GPT-3 model, comparing these methods to highlight how contextual factors influence hate speech. We identify both explicit hate speech and subtler forms of discriminatory language and microaggressions, illustrating the need for nuanced approaches to online toxicity. For example, a comment like "Sunday Gunday: Self-Defense" could be interpreted as either neutral or targeting specific groups, depending on its context. This work contributes to automated content moderation by enhancing the understanding and detection of targeting language, aiming to foster a more respectful online environment. The paper is structured as follows: Section 3 details the comprehensive annotation framework involving expert annotators, crowd annotators, and ChatGPT. Section 4 discusses the challenges and opportunities in annotating targeting language, including inter-annotator agreement and the limitations of ChatGPT.

Main contributions:

- **Comprehensive annotation framework:** Development and implementation of a detailed annotation framework using expert annotators, crowd annotators, and ChatGPT to label a diverse Reddit data set for inappropriately targeting language.

- **Comparative analysis:** A thorough comparative analysis of annotations from the experts, crowd, and ChatGPT, highlighting each approach's strengths and weaknesses in detecting implicit hate speech and microaggressions.

- **Identification of nuanced hate speech:** Detection of both explicit hate speech and subtler forms of discriminatory language, emphasizing how contextual factors impact hate speech manifestation.

- **Challenges and subjectivity in annotation:** Examination of the challenges and subjectivity involved in annotating targeting language, as well as issues of inter-annotator agreement and ChatGPT's difficulties with contextual nuances.

- **New target categories:** Identification of new targeting categories such as social belief, body image, addiction, and socioeconomic status, expanding the scope of hate speech detection research.

- **Addressing ChatGPT limitations** Highlighting the limitations of ChatGPT as an automated system in accurately identifying targeting language, emphasizing the need for advanced content moderation strategies that incorporate deeper contextual analysis and improve handling of nuanced language to enhance the effectiveness of AI-driven moderation tools.

## 2. Related Work

Recent research has seen a surge in efforts to detect and analyze targeting language on social media, driven by its importance in understanding online interactions and societal impacts. Various methods have been developed to identify and study such language.

For instance, [3] introduced a 5000-pair hate speech/counter narrative (HS/ CN) data set using a semi-automatic human-in-the-loop (HITL) approach with GPT-2 fine-tuning and iterative human review. This work offers diverse examples for NLP models combating online hate. Similarly, [4] created a data set from the Dutch LiLaH corpus, annotating 36,000 Facebook comments to identify hate speech types targeting groups like religion, gender, migrants, and the LGBT community. This data set focused exclusively on the prevalence of hateful metaphors. Although [3] and [4] focus on specific groups, such as race, religion, gender, migrants, and LGBT, their applicability to broader online hate speech contexts is limited.

The data set from [5] includes categorized tweets gathered from various sources such as Hatebase, Twitter hashtags, and public hate speech data sets. Human annotators classified these tweets into hate speech categories such as ethnicity or religion. The data set was compared with general Twitter data for context and used in diverse analyses exploring linguistic features, psychological dimensions, and semantic frames of hate speech. [5] also identifies explicit hate speech tweets using keyword-based filtering and manual annotation, collecting details about hate instigators and targets to study interaction patterns in online hate speech on Twitter. The Offensive Language Identification Dataset (OLID) [6] was designed to categorize offensive content in social media, particularly tweets. OLID employs a three-level hierarchical annotation schema: Level A distinguishes offensive (OFF) from non-offensive (NOT) tweets; Level B categorizes offensive tweets as targeted insults/threats (TIN) or untargeted offenses (UNT); Level C identifies the target of offensive language, classifying it into individual (IND), group (GRP), or other (OTH) targets. Data collection involved retrieving tweets using specific keywords and annotating the data set through crowdsourcing with experienced annotators. OLID is used for training and evaluating offensive language detection models, focusing on insults, threats, and targeted groups, but may overlook subtler forms of offensiveness. The keyword-based filtering methodology employed in data set creation by [5], [7], and [6] could miss nuanced instances of hate speech not containing specific keywords.

The data set used for abusive content classification by [8] consists of textual entries sourced from Reddit discussions accessed through the Pushshift API and Google's BigQuery. It employs social scientific concepts to detect and classify abuse across categories such as Neutral, Identity-directed abuse, Affiliation-directed abuse, Person-directed abuse, and Counter Speech. Trained annotators assigned labels to entries based on predefined categories, although inter-annotator agreement could be enhanced, particularly for challenging or ambiguous "edge case" content. The data set exhibits a skewed distribution towards the Neutral class, mirroring prevalent real-world patterns, with certain abusive categories appearing less frequently.

The reviewed studies have advanced our understanding of inappropriately targeting language online by creating specialized data sets and analytical methodologies. Each data set focuses on different aspects of hate speech, collectively revealing the complexities of online discourse. Building on this, our paper introduces a novel approach that broadens analysis to include more contextual factors and targets, enhancing hate speech detection comprehensively. The additional aspects of our work in comparison to the previous works are namely the integration of ChatGPT for annotation, a more detailed and specific annotation focus on targeting behaviors, offering insights about the comparative analysis of ChatGPT-generated annotations with human annotations, a more nuanced approach to target category identification, and discussing the subtleties and variations in the interpretation of online discourse.

## 3. Methodology

*3.1. Data Description*

We use a data set of English conversation threads from banned subreddits on Reddit, as created and selected by [9]. These subreddits were accessed via the Pushshift API and BigQuery. The data set includes 67,677 submissions and 1,168,546 comments, totaling 4,017,460 tokens. To ensure meaningful analysis, a structured approach inspired by [8] was employed to reconstruct conversation threads without overlap, starting each subthread with its initial comment. Subthreads were selected based on the following criteria: a minimum of 3 and a maximum of 17 comments, a token count range of 51 to 1,276 tokens per subthread, and a cap of 38 tokens per comment. Additionally, subthreads were categorized by their toxicity level using three lexicons. The top 400 most toxic subthreads and 98 non-toxic subthreads were filtered out to provide a robust representation of both high and low toxicity content. Out of 498 subthreads, 39 were selected as the gold data subset by [9]. This selection involved a stratified sampling method based on toxicity scores and comment counts, ensuring diverse representation. Text preprocessing steps included removing links and non-alphanumeric characters.

*3.2. Annotation Process*

Annotators identified inappropriate language targeting individuals or groups in comments or titles. They reviewed titles and comments, including associated context such as previous comments, title, and post text when available, with anonymized usernames guiding their assessments. Anonymization was performed according to [10]. Annotators exercised caution and discretion with potentially offensive content, adhering to the provided guidelines. Usernames in the examples presented in this paper are shown in brackets. Detailed annotation guidelines provided to annotators on the annotation platform, along with screenshots from the user interface that includes the instructions, can be found in Appendix C.

1. Annotators reviewed comments or titles to identify targeting towards individuals, groups, or broader categories. Items without targeting were labeled "not targeting," and annotators then proceeded with those that exhibited targeting behavior.

    **Example 3.1.** *Targeting: Furries should be in the same mental institutions as tran\*\*es. What in the f\*\*k happened to this country.*

2. Inside or outside of the conversation: Annotators determined if comments or titles targeted people within or outside the conversation thread, with Examples 3.2 and 3.3 highlighting these distinctions through underlined targets to clarify the concept.

**Example 3.2.** *Inside: go back to your fu\*\*king estro weed subs my dude.*

**Example 3.3.** *Outside: How else you gonna know what these retards are saying?*

A comment can target both a person inside the conversation and a person outside of the conversation, such as Example 3.4:

**Example 3.4.**

  *Context:*
   *[anon_8Wbs0]: I can't, he's right.*
   *[anon_8mUPN]: Hey I see your c\*ck fa\*\*ot a\*\* in the pic on the left.*
   *[anon_8Wbs0]: At least I'm not a racist loser, sweetie*

  *Comment: Being a racist is far better than being a c\*ck fa\*\*ot.*

In the context of Example 3.4, the speaker is both targeting the person they are addressing in the conversation and people with a particular sexual orientation in general outside of the conversation.

3. Identify target categories and target tokens: Annotators identified target categories, i.e., sexual orientation, gender, disability, age, race/ ethnicity/ nationality, religion, famous individual, political affiliation, and other target category (none of the aforementioned). They marked all relevant words in the comment or title pertaining to these categories, primarily focusing on nouns. Additional examples were given to assist annotators, as shown in Example 3.5, Example 3.6, and Example 3.7, where target tokens are underlined.

    **Example 3.5.** *Race/ ethnicity/ nationality: I don't want to know. It's usually dribble. I'll stick with more conservative black people who value their education.*

    **Example 3.6.** *Disability: What kind of a\*\*hole actually reports someone who is on their side politically? Fu\*\*king moron*

    **Example 3.7.** *Famous individual: The President having one of them clippers would be sweet. However, I think Trump deserves no less than a B-52.*

Furthermore, it is important to note that there could be more than one target category identified. For example, a comment can target both gender and disability, as in Example 3.8, where "c**t" refers to gender and "dullard" refers to disability.

**Example 3.8.** *Not only a c**t but a dullard too? Is there no beginning to your talent?*

Navigating challenges in achieving high inter-annotator agreement (IAA) due to subjective interpretation and nuanced instructions, we refined guidelines and implemented feedback mechanisms to enhance agreement through multiple rounds of discussions among the authors of the paper.

### 3.3. Crowd Annotation

All 498 subthreads were annotated by a crowd of five annotators. Annotator selection followed the method outlined by [11], involving pre-screening and post-screening to ensure quality and reliability. Detailed information about each annotator including their nationality, fluent languages, primary language, age, sex, ethnicity, country of residence, country of birth, employment status, and student status can be found in Appendix D. Crowd annotations were then adjudicated based on majority vote to create the AdjCrowd annotation set.

### 3.4. Expert Annotation

To evaluate the crowd annotation independently, three expert annotators (the authors) annotated the gold data. They underwent comprehensive training on the annotation platform for consistent guideline application. Annotations were adjudicated based on discussion and majority vote, forming the adjudicated annotation set referred to as "AdjExpert" for this study.

### 3.5. ChatGPT Annotation and Prompting

We integrated OpenAI's GPT-3 language model into our annotation process, using the text-davinci-003 engine to analyze and annotate textual data from 498 subthreads. ChatGPT was tasked with generating outputs for each step separately, building on its previous outputs. Prompt development, detailed in Appendix E, aimed for consistent annotation guidelines. We refined prompts through iterative analysis of conversations, incorporating feedback to enhance clarity and effectiveness in capturing nuanced aspects of inappropriate language.

### 3.5.1. Targeting or Not, Inside vs. Outside of the Conversation

To assess targets in conversations (comments and titles), we designed a prompt providing contextual details for each. This prompt guides the model to identify whether the target is within or outside the conversation during targeting. The model's varied responses are categorized to provide insights into the nature of targeting instances.

### 3.5.2. Target Category Annotation

In target category annotation, we tailored a prompt to identify specific targeted characteristics in comments or titles. ChatGPT responses in various formats were analyzed and categorized into predefined target categories based on the input from 3.5.1

### 3.5.3. Target Tokens Annotation

We created a prompt directing the model to focus on tokens related to identified target categories. These tokens, extracted from targeting comments or titles provided to ChatGPT along with their associated categories (as described in 3.5.3, offer additional context that enriches our analysis of targeting within the annotated data.

### 3.6. Evaluation and Statistical Analysis

We measured inter-annotator agreement by comparing levels of consistency among expert annotators, crowd annotators, and ChatGPT. Specifically, we analyzed the agreement between expert and ChatGPT annotations (AdjExpert vs. ChatGPT), between expert and crowd annotations (AdjExpert vs. AdjCrowd), as well as among expert annotations and among crowd annotations. For inter-annotator agreement analysis among both experts and crowd annotators, we computed average agreement scores across annotator pairs at both the comment level and subthread level. The original annotations were made at the comment level. Comment-level analysis considered annotations for each individual comment, whereas for subthread-level analysis, the comparison was based on aggregated comment labels, such that the whole subthread would be labelled as a particular category if at least one comment in that subthread was labelled as such. Target token comparison was limited to respective comments.

The upper section of Table 1 presents Cohen's Kappa scores on targeting at comment and subthread levels. The lower section focuses on agreement on target categories, inside/ outside of the conversation thread, and target tokens at comment and subthread levels given agreement on targeting. At the comment level, expert agreement on targeting achieved a moderate Cohen's Kappa score of 0.58, while the crowd scored lower at 0.36. This discrepancy reflects variations in expertise and experience, with experts demonstrating more consistent identification of targeting behaviors compared to the crowd. Subthread-level agreements improved (experts: 0.63, crowd: 0.54), indicating that aggregation at a broader level over multiple comments helps align interpretations. The higher agreement between expert and crowd annotations reflects a majority vote approach among crowd annotations versus expert labels, which tends to smooth out individual differences and align more closely with expert judgments. Conversely, the lower agreement among crowd annotators indicates greater variability in individual annotations, as it is based on averaged scores from varying annotator pairs. Therefore, this variability is due to the fact that the majority vote method for crowd annotations minimizes individual differences, enhancing apparent agreement with expert annotations, whereas direct comparison among crowd annotators reveals more inherent inconsistencies in their judgments due to their diverse interpretations and backgrounds. Agreement scores varied across categories, with some like "Age" and "Race" showing high consensus, while others like "Outside," "Gender", "Religion", and "Famous Individual" lower agreement, particularly among the crowd. "Other Target Category" exhibited consistently low agreement scores across all sources, indicating greater complexity and ambiguity.

| Kappa scores on targeting | | |
|---|---|---|
| | **Expert** | **Crowd** |
| **Comment-level** | 0.58 | 0.36 |
| **Subthread-level** | 0.63 | 0.54 |

| Kappa scores given agreement on being targeting | | | | |
|---|---|---|---|---|
| Category | Expert (Comment-level) | Crowd (Comment-level) | Expert (Subthread-level) | Crowd (Subthread-level) |
| Inside | 0.65 | 0.14 | 0.8 | 0.34 |
| Outside | 0.5 | 0.295 | 0.19 | 0.27 |
| Sexual Orientation | 0.68 | 0.56 | 0.66 | 0.6 |
| Gender | 0.57 | 0.42 | 0.37 | 0.41 |
| Disability | 0.62 | 0.556 | 0.72 | 0.54 |
| Age | 0.92 | 0.67 | 0.89 | 0.63 |
| Race | 0.73 | 0.58 | 0.67 | 0.61 |
| Religion | 0.605 | 0.75 | 0.59 | 0.73 |
| Famous Individual | 0.45 | 0.455 | 0.50 | 0.45 |
| Political Affiliation | 0.62 | 0.5 | 0.64 | 0.48 |
| Other Target Category | 0.11 | 0.26 | 0.16 | 0.24 |
| Target Tokens | 0.58 | 0.47 | - | - |

Table 1: Cohen's Kappa scores on targeting + Cohen's Kappa scores given agreement on being targeting (among the crowd, among the experts)

To further understand the inter-annotator agreement among experts, we analyzed the confusion matrices for targeting at both the comment and subthread levels, as presented in Tables 2 and 3. These matrices reveal that at the subthread level, annotators reached higher agreement compared to the comment level. This suggests that interpretations are more likely to align when considering aggregation at a broader level, thereby reducing the ambiguity associated with annotating individual comments. The high agreement at the subthread level indicates that, despite individual differences in annotating specific comments, annotators consistently converge on identifying the overall targeting behavior within a conversation thread.

Table 4 displays Cohen's Kappa scores comparing AdjExpert and AdjCrowd sets, as well as AdjExpert and ChatGPT-generated sets. ChatGPT-generated annotations show moderate agreement with AdjExpert annotations at both comment and subthread levels (0.4 and 0.37, respectively). In contrast, higher agreement exists between AdjExpert and AdjCrowd sets (0.58 at the comment level and 0.53 at the subthread level). This suggests closer alignment between AdjExpert and AdjCrowd compared to the ChatGPT-generated set.

|  | A1 vs. A2 | |
| --- | --- | --- |
|  | **Not targeting by A1** | **Targeting by A1** |
| Not targeting by A2 | 94 | 15 |
| Targeting by A2 | 34 | 105 |

|  | A1 vs. A3 | |
| --- | --- | --- |
|  | **Not targeting by A1** | **Targeting by A1** |
| Not targeting by A3 | 80 | 29 |
| Targeting by A3 | 26 | 113 |

|  | A2 vs. A3 | |
| --- | --- | --- |
|  | **Not targeting by A2** | **Targeting by A2** |
| Not targeting by A3 | 92 | 36 |
| Targeting by A3 | 14 | 106 |

Table 2: Confusion Matrices of Expert Annotators on the Annotation of Targeting at the Comment Level. Note: "A" is an abbreviation of "Annotator"

|  | A1 vs. A2 | |
| --- | --- | --- |
|  | **Not targeting by A1** | **Targeting by A1** |
| Not targeting by A2 | 2 | 2 |
| Targeting by A2 | 2 | 33 |

|  | A1 vs. A3 | |
| --- | --- | --- |
|  | **Not targeting by A1** | **Targeting by A1** |
| Not targeting by A3 | 3 | 1 |
| Targeting by A3 | 1 | 34 |

|  | A2 vs. A3 | |
| --- | --- | --- |
|  | **Not targeting by A2** | **Targeting by A2** |
| Not targeting by A3 | 3 | 1 |
| Targeting by A3 | 1 | 34 |

Table 3: Confusion Matrices of Expert Annotators on the Annotation of Targeting at the Subthread Level. Note: "A" is an abbreviation of "Annotator"

Some variation can be observed in agreement levels across different categories and annotation sources. For example, "Sexual Orientation" exhibits relatively high agreement between the ChatGPT-generated set and AdjExpert set (com-ment: 0.66, subthread: 0.87) as well as between AdjCrowd and AdjExpert set (comment: 0.74, subthread: 0.75). Similarly, "Religion" shows high agreement between the ChatGPT-generated set and AdjExpert set (comment: 0.66, subthread: 0.48) and between AdjCrowd and AdjExpert set (comment: 0.66, subthread: 0.48). Conversely, between the ChatGPT-generated set and the AdjExpert set, "Famous Individual" displays very low agreement (comment: 0, subthread: -0.05), indicating challenges in consistently identifying targeting in these contexts. As for the identification of targeting "Famous Individual", ChatGPT was able to identify only one case as such although identified by experts as not targeting at all (Example 3.9).

| Kappa scores on targeting | | |
|---|---|---|
| | **AdjExpert vs. ChatGPT** | **AdjExpert vs. AdjCrowd** |
| **Comment-level** | 0.4 | 0.58 |
| **Subthread-level** | 0.37 | 0.53 |

| Kappa scores given agreement on being targeting | | | | |
|---|---|---|---|---|
| Category | ChatGPT vs. Adj.Expert (Comment-level) | AdjExpert vs. AdjCrowd (Comment-level) | ChatGPT vs. AdjExpert (Subthread-level) | AdjExpert vs. AdjCrowd (Subthread-level) |
| Inside | 0.09 | 0.55 | 0.37 | 0.70 |
| Outside | 0.08 | 0.65 | 0 | 0.46 |
| Sexual Orientation | 0.66 | 0.74 | 0.87 | 0.75 |
| Gender | 0.06 | 0.52 | 0.48 | 0.66 |
| Disability | -0.015 | 0.49 | 0.12 | 0.47 |
| Age | 0.34 | 0.74 | 0.545 | 0.63 |
| Race | 0.55 | 0.62 | 0.59 | 0.8 |
| Religion | 0.66 | 0.66 | 0.48 | 0.48 |
| Famous Individual | 0 | 0.695 | -0.05 | 0.77 |
| Political Affiliation | 0.49 | 0.7 | 0.53 | 0.47 |
| Other Target Category | 0 | 0.14 | 0.02 | 0.18 |
| Target Tokens | 0.68 | 0.57 | - | - |

Table 4: Cohen's Kappa scores on targeting + Cohen's Kappa scores given agreement on being targeting (AdjExpert vs. AdjCrowd, ChatGPT vs. AdjExpert)

**Example 3.9.** *If you think Demi Lovato is ugly, you have probably never had sex with a human being before. Not even a gay guy would think she's ugly.*

Example 3.10 is an example of "Famous Individual" not detected correctly, where ChatGPT labeled it as targeting "Other Target Category", while experts labeled it as targeting "Famous Individual". In this example, the complexity of language and indirect reference to President Donald J. Trump might have challenged ChatGPT's ability to connect the comment directly to a famous individual.

**Example 3.10.**

*Context:*

*Title: WATCH PARTY: USA-NK Summit Coverage & Leader Arrivals*

*[anon_2Uwug]: I ignore them. Make them irrelevant*

*Comment: That's all well and good until he starts yelling stupid shit during a meeting of this magnitude*

Overall, crowd annotations demonstrate higher agreement with expert annotations compared to ChatGPT-generated annotations. We also analyzed the confusion matrices to evaluate the performance and agreement levels between ChatGPT-generated annotations and AdjExpert, as well as between AdjCrowd and AdjExpert, at both the comment and subthread levels. Figure 1 indicates that ChatGPT tends to over-identify comments as targeting, demonstrating higher sensitivity but lower specificity compared to expert annotations. Figure 2 suggests better alignment between ChatGPT and AdjExpert annotations at the subthread level, with ChatGPT showing higher precision in identifying targeting behavior when considering the aggregation at a broader level. The discrepancy between the higher alignment indicated by the confusion matrices and the slightly lower Cohen's Kappa scores at the subthread level compared to the comment level can be attributed to several factors. Cohen's Kappa adjusts for chance agreement, which can differ significantly depending on the prevalence of targeting labels and the number of instances being evaluated. At the comment level, with more data points, the absolute number of agreements and disagreements is higher, affecting the expected chance agreement. Additionally, at the subthread level, the aggregation of comments might reduce the penalization for misclassifications, but it still influences the overall Kappa calculation. Consequently, the apparent better raw alignment at the subthread level does not translate directly into a higher Kappa score, reflecting the nuanced nature of chance-adjusted agreement metrics. Figure 3 indicates that while there is good agreement between crowd and expert annotations, there are still notable discrepancies, likely due to varying levels of understanding and interpretation of targeting criteria among crowd annotators. Figure 4 suggests that aggregation at a broader level helps align crowd annotations more closely with expert annotations, reducing individual variations.

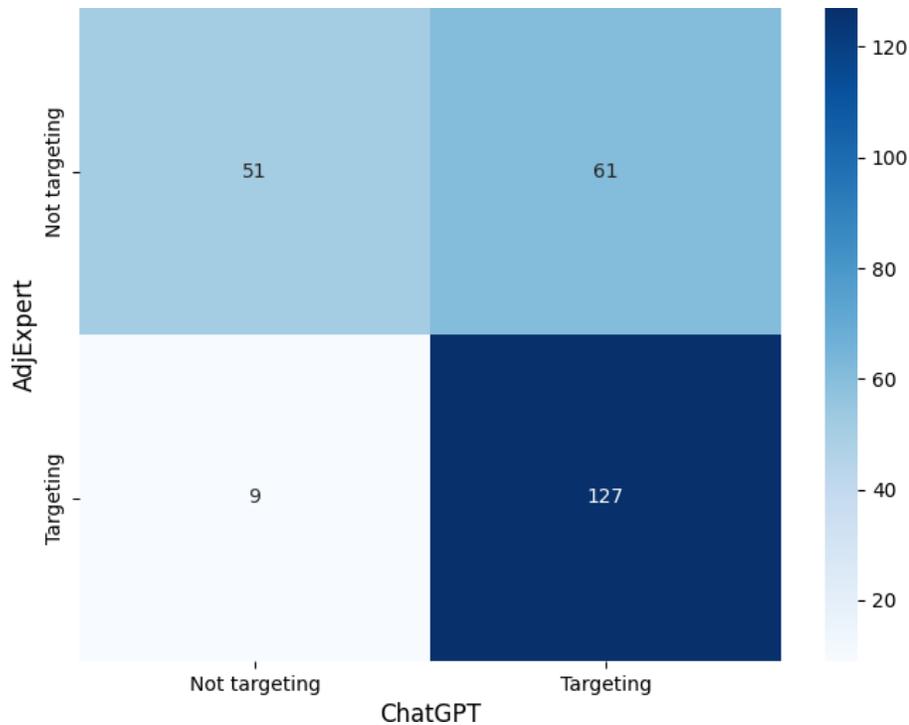

Figure 1: Confusion matrix of ChatGPT vs. AdjExpert on the annotation of targeting at the comment level

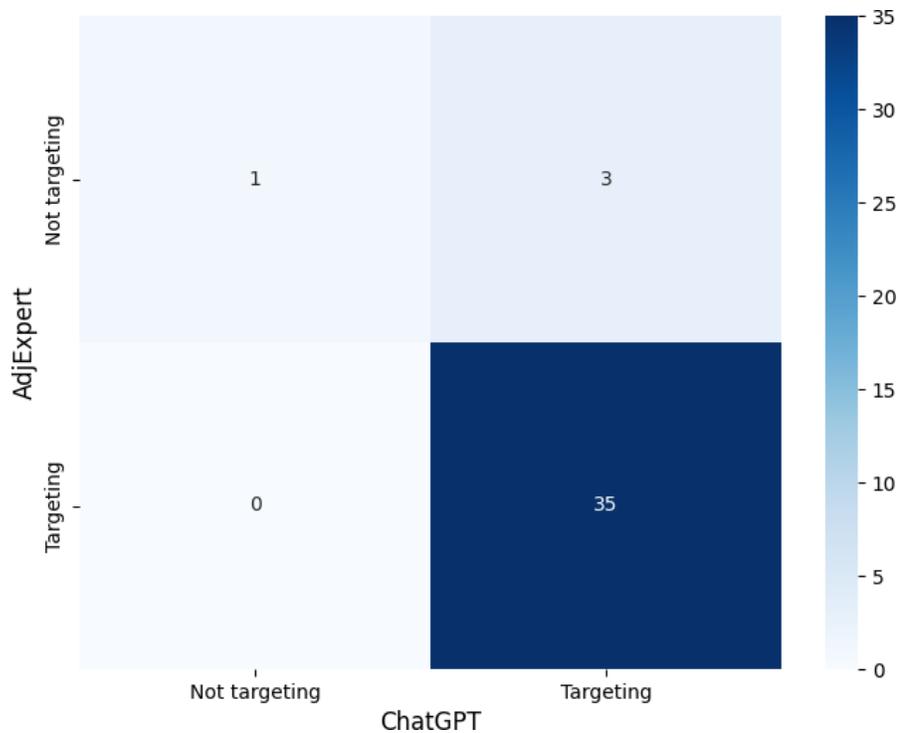

Figure 2: Confusion matrix of ChatGPT vs. AdjExpert on the annotation of targeting at the subthread Level

Table 5 summarizes the number of annotations for targeting, per target category, and inside vs. outside conversation threads, at comment and subthread levels, from expert annotators, crowd annotators, and ChatGPT. To ensure comparability, expert and crowd counts were normalized by dividing total annotations by the number of annotators. Expert annotators averaged 133.67 targeting comments annotated, slightly lower than the crowd's 141.4. ChatGPT generated notably more, totaling 188 targeting annotations at the comment level. At the subthread level, experts averaged 35 targeting annotations, the crowd 36, while ChatGPT annotated 38. Categories like "Religion" and "Famous Individual" received fewer annotations across all sources. ChatGPT was more over-identified targeting language compared to the experts and crowd but struggled more to discern targeting inside the conversation in comparison to the human annotators.

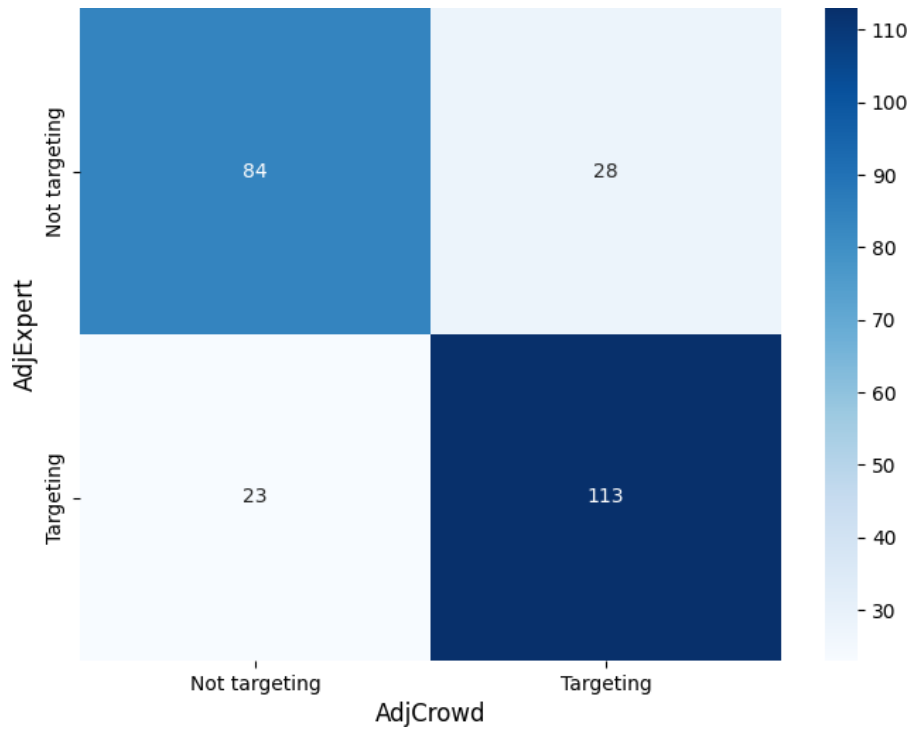

Figure 3: Confusion matrix of AdjExpert vs. AdjCrowd on the annotation of targeting at the comment level

As already demonstrated in this section, the moderate agreement between expert and crowd annotators indicates consistent identification of targeting behaviors. Section 4 further explores discrepancies among expert annotators. While we know that ChatGPT handles large data volumes at a lower cost and faster compared to humans, lower agreement scores compared to humans reveal challenges with context and subtle targeting behaviors, including cultural nuances. This emphasizes the importance of refining AI-driven content

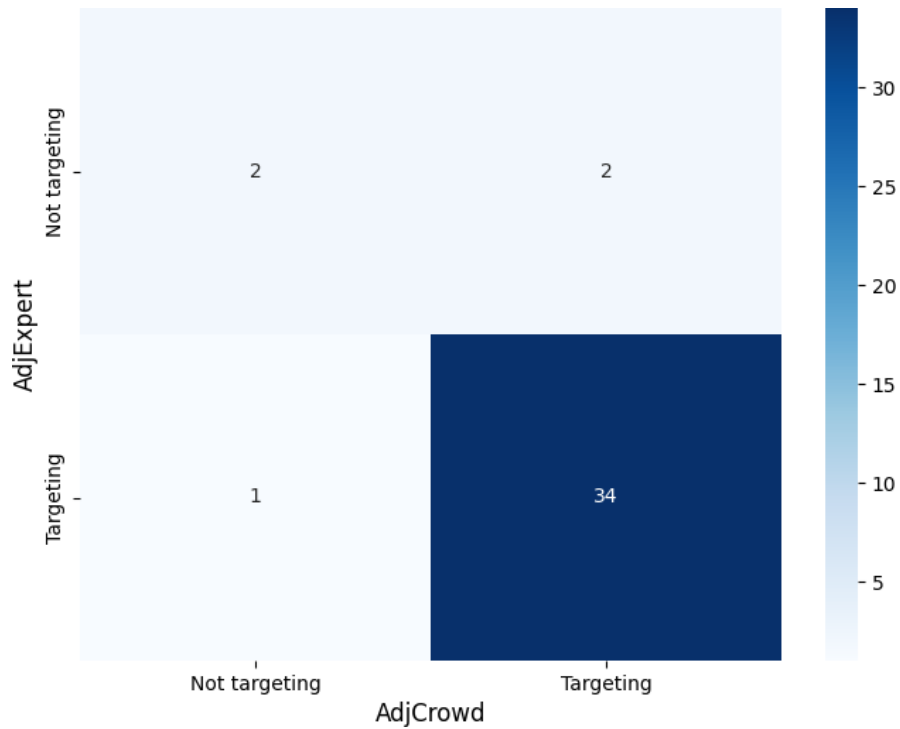

Figure 4: Confusion matrix of AdjExpert vs. AdjCrowd on the annotation of targeting at the subthread level

moderation for better contextual comprehension and cultural sensitivity.

## 4. Analysis and Discussion

### 4.1. Disagreements Among the Experts

We concentrated on cases where expert annotators disagreed on comment-level targeting annotations, revealing discrepancies among annotators where one deviated from the consensus. We categorized these disagreements to explore their nuances and potential origins, presenting an error analysis with specific examples for each category.

1. Different criteria for inappropriate language (targeting but not necessarily targeting inappropriately):

| Category | Comment-level | | | Subthread-level | | |
|---|---|---|---|---|---|---|
| | Expert | Crowd | ChatGPT | Expert | Crowd | ChatGPT |
| Targeting | 133.67 | 141.4 | 188 | 35 | 36 | 38 |
| Inside | 70.33 | 78 | 9 | 20 | 22.4 | 6 |
| Outside | 59.67 | 61.8 | 179 | 28.33 | 28.2 | 38 |
| Sexual Orientation | 32.67 | 24 | 33 | 11.33 | 14.6 | 12 |
| Gender | 24.33 | 19.2 | 25 | 14.67 | 11.8 | 9 |
| Disability | 20.67 | 12.2 | 1 | 10.67 | 6.6 | 1 |
| Age | 4.33 | 5 | 8 | 3.33 | 3.8 | 7 |
| Race | 16 | 19.8 | 32 | 6 | 8.4 | 10 |
| Religion | 1.67 | 4.2 | 3 | 1.67 | 3.2 | 3 |
| Famous Individual | 9.33 | 10.4 | 1 | 5.33 | 8 | 1 |
| Political Affiliation | 13.67 | 22 | 19 | 8.33 | 12.6 | 8 |
| Other Target Category | 26.33 | 50.8 | 69 | 15 | 22.4 | 30 |
| Target Tokens | 197.67 | 389.4 | 303 | - | - | - |

Table 5: Number of annotations

**Example 4.1.** *Let me give you all a hint. Hillary's statement the other day was not a spontaneous, meaningless slip up*

Annotators may argue that the statement does not always imply inappropriately targeting. They suggest it might be analyzing or discussing a public figure's actions, which might not always amount to inappropriately targeting. This discrepancy arises from differing interpretations of what qualifies as inappropriately targeting. In Example 4.1, while some see the reference to the famous individual as targeting inappropriately, others take into account nuanced factors like tone and context before deeming it as such, considering it as not inappropriate and only targeting.

2. Inherent complexity/ ambiguity

**Example 4.2.** *Sunday Gunday: Self-Defense*

The varied interpretation of Example 4.2 sparks disagreement among annotators. It may target specific groups, like Sunday gun enthusiasts promoting self-defense, or describe activities without clearly targeting, highlighting complexity and ambiguity, making unanimous agreement challenging.

3. Contextual understanding

**Example 4.3.**

*Context:*

> *Title: Me during fag pride month*
>
> *[anon Khiib]: Why the he** do You Care if guys like d**k instead of pu**y. Relax dude*
>
> *[anon 4OsvY]: They're degenerate f**ks.*
>
> *[anon BbKt2]: Dann, how long it take you to write that thesis?*
>
> *[anon yDQWE]: you don't need a thesis to see something so simple*
>
> *[anon BbKt2]: Then you should be able to explain easily: Why does it matter if a guy likes d**k instead of p***y?*
>
> *[anon EzrXS]: How is it an accomplishment?*
>
> *Comment: Its not? Why is it degeneracy?*

Disagreement over Example 4.3 stems from differing interpretations of "degeneracy." One annotator might view it as targeting sexual orientation, while another might see it as critiquing behavior or societal norms. This highlights the need to consider broader context and language nuances, urging discretion in annotator judgments, especially when interpretations vary with conversational context.

4. Subjective interpretation on the concept of targeting

   **Example 4.4.** *Me during fag pride month*

   The disagreement on Example 4.4 stems from differing interpretations of targeting. Some might view the term "fag" as derogatory, targeting individuals, especially during "pride month", while others might see it as self-referential or reclaimed, celebrating LGBTQ+ identities.

5. Insufficient contextual information

   **Example 4.5.**

   > *Context: Interesting thing to say when you're objectively below average looking*
   >
   > *Comment: Cause you can get called a rapist and get convicted if the girl regrets sleeping with you.*

   The lack of explicit context or cues in Example 4.5 makes it difficult for annotators to accurately identify the intended target. This highlights the importance of contextual information for precise annotation and interpretation.

6. Different interpretation of sarcasm

   **Example 4.6.**

> *Context:*
>
> *Title: SO TRUE!!*
>
> *[anon_py8O2]: you're a c**t*
>
> *[anon_aWz24]: You're 14*
>
> *[anon_ozqK8]: not only a c**t but a dullard too? is their no beginning to your talent?*
>
> *[anon_FTutK]: 12, then.*
>
> *[anon_ETMDg]: you weirdo liberal folks sure have an unhealthy interest in talking with young people, are you trying to groom me?*
>
> *Comment: Well, you tried.*

Disagreements often arise over interpreting sarcasm in comments. For instance, in Example 4.6, one annotator might perceive the comment as sarcastic or dismissive, suggesting the person's effort was futile, while another might view it as supportive or encouraging, acknowledging effort despite the outcome. These differing interpretations highlight the subjective nature of assessing language nuances, especially when tone indicators are not clear.

Additionally, we analyzed cases where expert annotators unanimously identified the "other target category" within comments and classified these instances into new categories to grasp their nuances. The new categories discovered in the data are social belief, body image, addiction, and socioeconomic status.

### 4.2. Analysis of ChatGPT's Identification of Targeting Language

Our examination of ChatGPT's annotations versus expert assessments on targeting language revealed 37 discrepancies where ChatGPT identifies comments as targeting, but experts unanimously disagree. Out of all gold data, ChatGPT marks 75% of cases as targeting, while experts (based on the majority vote) deem only 55% as such. A majority of cases flagged by ChatGPT as targeting are actually neutral according to expert assessments (see Example 4.7). Additionally, some comments contain toxic language but do not meet the criteria for targeting. For instance, comments deemed inappropriate or containing slurs do not necessarily imply targeting language directed at an individual or group (see Example 4.8). The findings suggest that ChatGPT's classification leans towards over-identification of targeting language, which can result in inaccurate annotations and potentially misleading interpretations. The system's approach lacks nuanced understanding of contextual cues and the subtleties of language that human experts can readily discern. One notable observation is that ChatGPT's misclassifications often involve comments where humor, sarcasm, or colloquialisms are present. For instance, light-hearted jokes or playful banter may be mistakenly categorized as targeting due to ChatGPT's limited ability to grasp nuanced language (see Example 4.9 and Example 4.10). Another critical insight stems from ChatGPT's handling of comment structure. Notably, ChatGPT appears to interpret titles and first comments as non-targeting. This observation suggests that ChatGPT's comprehension of targeting versus non-targeting language may be influenced by structural elements.

**Example 4.7.** *747-8 is the newest generation. The current AF1s are 747-200s.*

**Example 4.8.** *Why the hell do You Care if guys like d\*\*k instead of pu\*\*y. Relax dude*

**Example 4.9.** *LOL rememeber how triggered they were when he got the nomination?*

**Example 4.10.** *Ouch, epic burn, my friend. How would I ever recover*

## 5. Conclusion

This paper explores hate speech detection in online conversations through crowd annotation, expert annotation, and ChatGPT integration. Our analysis uncovered explicit hate speech as well as subtler discriminatory language and microaggressions. This study contributes a benchmark data set for analyzing targeting language and evaluating automated detection systems, facilitating research into harmful online communication and supporting the development of improved content moderation strategies. We observed moderate agreement between expert and crowd annotators, yet challenges remain in consistently identifying targeting behaviors, particularly in nuanced contexts such as distinguishing targets within or outside conversations and assessing subjective targeting language. ChatGPT showed that it tended to over-label, identifying more instances of targeting language compared to human annotators. This highlights the need for further refinement in automated moderation systems. By categorizing specific target categories like gender, political affiliation, and sexual orientation, this study enhances our understanding of harmful online communication and informs the development of more effective moderation strategies to promote inclusive and respectful online environments. Future research should focus on refining annotation guidelines, enhancing contextual NLP model understanding, and developing scalable moderation methods. Additionally, efforts should be made to expand the data set to include a broader range of targeting instances and improve its utility as a benchmarking resource.

## 6. Limitations

While our study offers valuable insights into detecting and analyzing inappropriately targeting language in online platforms, certain limitations must be considered. Firstly, the size and diversity of the data set may limit the generalizability of our findings, despite efforts to collect a varied range of conversation threads from Reddit. We acknowledge that the focus on banned subreddits may not fully capture the diversity of online hate speech across different platforms and communities. Additionally, inherent biases in the annotation process, influenced by annotators' subjective interpretations and contextual understanding, may affect the reliability of labels assigned to comments. The predefined target categories may not fully encompass the spectrum of inappropriately targeting language, and emerging forms of online harassment may not be adequately captured. While ChatGPT assisted in the annotation process, its performance varies depending on text complexity and ambiguity, introducing errors. Inter-annotator agreement challenges and the limited generalizability of findings to other platforms further underscore the need to interpret results cautiously and guide future research directions effectively. Finally, our annotation and analysis is limited to English. This should be expanded to more diverse languages and cultures.

## 7. Acknowledgments

This research was supported by Huawei Finland through the DreamsLab project. All content represented the opinions of the authors, which were not necessarily shared or endorsed by their respective employers and/ or sponsors.

## Appendix A. Annotation Platform and Annotator Recruitment

The annotation task was designed and implemented using the LingoTURK platform [12], which provided a user-friendly interface for annotators to complete the tasks efficiently. Annotators were recruited through the Prolific platform [13], which allowed for the selection of a diverse pool of participants to ensure comprehensive annotation coverage. This combination of annotation platform and recruitment strategy facilitated the collection of annotations from a broad range of annotators, contributing to the reliability and validity of the data set.

## Appendix B. Ethical Considerations

This study adhered to ethical guidelines, ensuring the privacy and anonymity of Reddit users and annotators. The annotation task contained content warnings, and participants were informed about the nature of the task before engaging in the annotation process.

**Appendix C. Annotation Guidelines and User Interface Screenshots**

Figure C.5 demonstrates the instructions page shown to the annotators on the annotation platform. Figure C.9 is an example of the way the comments in a conversation were shown to the annotators to be annotated, starting from the title text (Figure C.6), to comment 1 (Figure C.7) and (Figure C.8). The real conversation is longer than what presented here up to the second comment.

**Appendix D. Crowd Annotators' Information**

| Fluent Languages | Percentage |
| --- | --- |
| English | 99.75% |
| Chinese | 0.74% |
| Spanish | 1.48% |
| French | 1.73% |
| Afrikaans | 0.25% |
| German | 1.23% |
| Italian | 0.25% |
| Turkish | 0.25% |
| Dutch | 0.25% |
| Japanese | 0.25% |
| Hindi | 0.49% |
| CONSENT_REVOKED | 0.25% |
| Thai | 0.49% |
| Gaelic | 0.25% |
| Romanian | 0.25% |
| Urdu | 0.49% |
| Bengali | 0.25% |
| Greek | 0.25% |

Table D.6: Statistics for fluent languages. Note that the percentages are not mutually exclusive, as annotators may speak more than one language.

Tables D.7, D.8, D.9, D.10, D.11, D.12, D.13, D.14, and D.15 provide statistical information of the crowd annotators' backgrounds.

## Instructions

Please be aware that this task contains sentences that are not appropriate. Therefore, we would like to ask you not to participate if you do not feel comfortable being exposed to such content.

### Step 1: Preliminary Assessment of Targeting

You will review comments/titles posted on Reddit that may contain inappropriate language. Your task is to identify if they target a specific person or group of people. You should first read the post text and follow these steps for each comment/title:

1. Read the comment/title and assess whether it targets anyone.
2. If the comment/title does not target someone, mark the comment/title as "not targeting". If the comment/title targets someone, proceed to Step 2.

### Step 2: Determine the Targets

In this step, you will decide whether the comment/title is targeting someone inside or outside of the conversation.

**Examples:**

```
inside: Sit the fuck down, you colossal piece of shit.

inside: Stupid me did not press the record button.

outside: They should all be ashamed of themselves.
```

### Steps 3-4: Identify Target Categories and Their Reference Words

Now determine if a person or group is targeted based on certain characteristics known as target categories by identifying them. Mark all the words in the comment/title that refer to the identified target categories.

**Examples:**

1. sexual orientation: Gay men refuse to fuck like any other person.

   reference words: Gay, men

   gender: This person is a cisgender man pretending to be a woman to sell his shitty bl

   reference words: cisgender, man

   disability: Why are you just sitting there like a mentally paralyzed person?

   reference words: mentally, paralyzed

   age: He still acts like a child.

   reference words: child

   race, ethnicity, or nationality: I don't give a fuck if you are Asian.

   reference words: Asian

   religion: I literally hate Christians so much.

   reference words: Christians

   famous individual: How embarrassing Serena Williams was at the US open when she lost

   reference words: Serena, Williams

   political affiliation: Good morning to everyone who isn't a liberal fuck!

   reference words: liberal

Note: If a comment/title targets multiple categories, mark all reference words for all categories.

**Example:**

1. sexual orientation, famous individual: Fuck Pitbull and Ne-Yo gay ass

   reference words: Pitbull, Ne-Yo, gay

In case a target category isn't listed, mark it as "other target category."

Please be aware that your answers will be analyzed based on and compared with those of other annotators and annotators with low consistency may be excluded from future versions of this experiment. Therefore, we will not be able to pay those who do not change their behavior despite warnings.

This task is part of the MIAMI project coordinated by the Faculty of Humanities, the Computational Linguistics and Text Mining Lab (CLTL) at the Vrije Universiteit Amsterdam, aiming at identifying toxicity in text. Your contribution is really appreciated as it helps fight hate crimes on social media.

Your answers will be used for scientific purposes. No personal information will be stored.

If you have any questions or feedback, please contact **Baran Barbarestani** via **b.barbarestani@vu.nl**.

Next

Figure C.5: Instructions

| Primary Language | Percentage |
|---|---|
| English | 99.75% |
| Turkish | 0.25% |
| Other | 0.25% |
| CONSENT_REVOKED | 0.25% |
| Thai | 0.49% |
| Urdu | 0.49% |
| Esperanto | 0.25% |

Table D.7: Statistics for primary language. Note that the percentages are not mutually exclusive, as annotators may report more than one primary language.

| Nationality | Percentage |
|---|---|
| United Kingdom | 99.75% |
| CONSENT$_R$EV OKED | 0.25% |

Table D.8: Statistics for nationality

| Age Range | Percentage |
|---|---|
| 31-40 | 31.11% |
| 51-60 | 10.12% |
| 41-50 | 16.79% |
| 21-30 | 34.07% |
| 11-20 | 1.98% |
| 61-70 | 4.94% |
| 120-130 | 0.25% |
| 71-80 | 0.74% |

Table D.9: Statistics for age

| Sex | Percentage |
|---|---|
| Female | 66.91% |
| Male | 32.84% |
| CONSENT_REVOKED | 0.25% |

Table D.10: Statistics for sex

| Ethnicity | Percentage |
|---|---|
| White | 85.68% |
| Asian | 4.20% |
| Mixed | 6.91% |
| Black | 2.72% |
| CONSENT_REVOKED | 0.25% |
| Other | 0.25% |

Table D.11: Statistics for ethnicity

| Country of Birth | Percentage |
|---|---|
| United Kingdom | 99.26% |
| CONSENT_REVOKED | 0.25% |
| United States | 0.25% |
| Canada | 0.25% |

Table D.12: Statistics for country of birth

| Student Status | Percentage |
|---|---|
| No | 72.84% |
| DATA_EXPIRED | 16.79% |
| Yes | 10.12% |
| CONSENT_REVOKED | 0.25% |

Table D.13: Statistics for student status

# Appendix E. ChatGPT prompts

| Employment Status | Percentage |
|---|---|
| DATA_EXPIRED | 20.49% |
| Full-Time | 40.25% |
| Part-Time | 14.81% |
| Not in paid work (e.g. homemaker) | 14.32% |
| Retired or disabled | 14.32% |
| Other | 3.95% |
| Due to start a new job within the next month | 1.48% |
| Unemployed (and job seeking) | 4.44% |
| CONSENT_REVOKED | 0.25% |

Table D.14: Statistics for employment status

| Country of Residence | Percentage |
|---|---|
| United Kingdom | 94.81% |
| Australia | 0.74% |
| Spain | 0.49% |
| South Africa | 0.49% |
| Netherlands | 0.25% |
| Ireland | 1.23% |
| United States | 0.25% |
| CONSENT_REVOKED | 0.25% |
| Japan | 0.74% |
| Canada | 0.25% |
| France | 0.25% |
| New Zealand | 0.25% |

Table D.15: Statistics for country of residence

Figure C.6: *
Title text

Figure C.7: *
Comment 1

Figure C.8: *
Comment 2

Figure C.9: User interface

> You are a language model that helps annotate inappropriate words and harmful messages in comments. Given a CSV file, First, you should read the following columns per row: "title_text", "comment", "context", "user_id", and "post_text". "context" refers to the previous comments. You should read each "title_text", "post_text", and "context". Then, you should read "comment". If "comment" is not "-", you should produce your response based on "comment". If "comment" is "-", you should produce your response based on "title text".
> Instructions
> You will review comments/titles posted on Reddit that may contain inappropriate language. Your task is to identify if they target a specific person or group of people. You should first read the post text. Read the comment/title and assess whether it targets anyone. If the comment/title does not target someone, mark the comment/title as "not targeting". If the comment/title targets someone, you should decide whether the comment/title is targeting a person inside the conversation thread or outside of the conversation thread. Mark targeting a person inside the conversation thread as I. Mark targeting a person outside of the conversation thread as O. Mark not targeting as N
> Examples:
> I: Sit the f**k down, you colossal piece of s**t.
> I: Stupid me did not press the record button.
> O: They should all be ashamed of themselves.
> N: The sky is clear and the sun is shining brightly today.
> N: This f**king life is crazy!

Figure E.10: Prompt for ChatGPT annotation of targeting and inside vs. outside of the conversation

> You are a language model that helps annotate inappropriate words and harmful messages in comments. Given a CSV file, First, you should read the following columns per row: "title_text", "comment", "context", "user_id", and "post_text". "context" refers to the previous comments. You should read each "title text", "post text", and "context". Then, you should read "comment". If "comment" is not "-", you should produce your response based on "comment". If "comment" is "-", you should produce your response based on "title_text".
> Instructions
> You will review comments/titles posted on Reddit that may contain inappropriate language. Your task is to identify if they target a specific person or group of people. You should first read the post text. Read the comment/title and assess whether it targets anyone. If the comment/title does not target someone, mark the comment/title as "not targeting". If the comment/title targets someone, you should decide whether the comment/title is targeting a person inside the conversation thread or outside of the conversation thread. Mark targeting a person inside the conversation thread as I. Mark targeting a person outside of the conversation thread as O. Mark not targeting as N
> Examples:
> sexual orientation: Gay men refuse to f**k like any other person.
> gender: This person is a cisgender man pretending to be a woman to sell his sh**ty blog.
> disability: Why are you just sitting there like a mentally paralyzed person?
> age: He still acts like a child.
> race, ethnicity, or nationality: I don't give a f**k if you are Asian.
> religion: I literally hate Christians so much.
> famous individual: How embarrassing Serena Williams was at the US open when she lost to Osaka.
> political affiliation: Good morning to everyone who isn't a liberal f**k!
> other target category: We should send all refugees to concentration camps.
> no target category: I hate you!
> Note: If a comment/title targets multiple categories, you should mark all target categories.
> Examples:
> sexual orientation, famous individual: F**k Pitbull and Ne-Yo gay a**
> other target category, religion: Unemployed people, particularly Muslims, are like parasites as they live off social welfare.

Figure E.11: Prompt for ChatGPT annotation of target categories

> You are a language model that helps annotate inappropriate words and harmful messages in comments. You should read each "Title", "Post Text", "Context", "User ID", "Comment", "Text", and "Target categories". Then, you should produce your response based on "Text". "Context" refers to the previous comments.
>
> Instructions
>
> You will review comments/titles posted on Reddit that target a person or group of people inappropriately due to certain characteristics known as target categories. Your task is to identify the tokens that make reference to those target categories.
>
> List of target categories: 1. Sexual orientation 2. Gender 3. Disability 4. Age 5. Race, ethnicity, or nationality 6. Religion 7. Famous individual 8. Political affiliation
>
> Examples:
>
> Text: Gay men refuse to f**k like any other person.
>
> Target categories: Sexual orientation
>
> Target tokens: Gay, men
>
> Text: This person is a cisgender man pretending to be a woman to sell his sh**ty blog.
>
> Target categories: Gender
>
> Target tokens: cisgender, man

Figure E.12: Prompt for ChatGPT annotation of target tokens (Part 1)

> Text: Why are you just sitting there like a mentally paralyzed person?
> Target categories: Disability
> Target tokens: mentally, paralyzed
> Text: He still acts like a child.
> Target categories: Age
> Target tokens: child
> Text: I don't give a f**k if you are Asian.
> Target categories: Race, ethnicity, or nationality
> Target tokens: Asian
> Text: I literally hate Christians so much.
> Target categories: Religion
> Target tokens: Christians
> Text: How embarrassing Serena Williams was at the US open when she lost to Osaka.
> Target categories: Famous individual
> Target tokens: Serena, Williams

Figure E.13: Prompt for ChatGPT annotation of target tokens (Part 2)

Figures E.10, E.11, E.12, and E.13 demonstrate the designed prompts for Chat-GPT. Please note that the slurs and inappropriate words that are obfuscated here are not so in the real prompts.